\documentclass{article}
\usepackage{ijcai21}

\usepackage{times}
\usepackage{soul}
\usepackage{url}
\usepackage[bookmarksnumbered=true]{hyperref}
\usepackage[utf8]{inputenc}
\usepackage[small]{caption}
\usepackage{graphicx}
\usepackage{amsmath}
\usepackage{amsthm}
\usepackage{booktabs}
\usepackage{algorithm}
\usepackage{algorithmic}
\usepackage{amsfonts}
\usepackage{bm}
\usepackage{tabularx}
\usepackage[switch]{lineno}
\newtheorem{proposition}{Proposition}
\newcommand{\bhline}[1]{\noalign{\hrule height #1}}
\newcolumntype{C}{>{\centering\arraybackslash}X}
\newcolumntype{L}{>{\raggedright\arraybackslash}X}
\newcolumntype{R}{>{\raggedleft\arraybackslash}X}
\newcommand{\myfootnote}[1]{
	\renewcommand{\thefootnote}{}
	\footnotetext{\hspace{-16.5pt}#1}
	\renewcommand{\thefootnote}{\arabic{footnote}}
}

\title{Decomposable-Net: Scalable Low-Rank Compression for Neural Networks}

\author{
Atsushi Yaguchi$^1$ \and
Taiji Suzuki$^{2,3}$ \and
Shuhei Nitta$^1$ \and
Yukinobu Sakata$^1$ \And
Akiyuki Tanizawa$^1$ \\
\affiliations
$^1$Toshiba Corporation, Japan \\
$^2$The University of Tokyo, Japan \\
$^3$RIKEN Center for Advanced Intelligence Project, Japan \\
\emails
\{atsushi.yaguchi,shuhei.nitta,yuki.sakata,akiyuki.tanizawa\}@toshiba.co.jp, 
taiji@mist.i.u-tokyo.ac.jp
}
\begin{document}

\maketitle

\begin{abstract}
Compressing DNNs is important for the real-world applications operating on resource-constrained devices. 
However, we typically observe drastic performance deterioration when changing model size after training is completed. 
Therefore, retraining is required to resume the performance of the compressed models suitable for different devices. 
In this paper, we propose Decomposable-Net (the network decomposable in any size), 
which allows flexible changes to model size without retraining. 
We decompose weight matrices in the DNNs via singular value decomposition and adjust ranks according to the target model size. 
Unlike the existing low-rank compression methods that specialize the model to a fixed size, 
we propose a novel backpropagation scheme that jointly minimizes losses for both of full- and low-rank networks. 
This enables not only to maintain the performance of a full-rank network {\it without retraining} but also to improve low-rank networks in multiple sizes. 
Additionally, we introduce a simple criterion for rank selection that effectively suppresses approximation error. 
In experiments on the ImageNet classification task, Decomposable-Net yields superior accuracy in a wide range of model sizes. 
In particular, 
Decomposable-Net achieves the top-1 accuracy of $73.2\%$ with $0.27\times$MACs with ResNet-50, 
compared to Tucker decomposition ($67.4\% / 0.30\times$), Trained Rank Pruning ($70.6\% / 0.28\times$), and universally slimmable networks ($71.4\% / 0.26\times$).
\end{abstract}

\myfootnote{Copyright \copyright 2021 International Joint Conference on Artificial Intelligence (IJCAI). All rights reserved.}
\myfootnote{The code is available at {\scriptsize \texttt{\url{https://github.com/ygcats/Scalable-Low-Rank-Compression-for-Neural-Networks}}}.}

%-------------------------------------------------------------------------------------------------------------------------------------------------------------------------------------
\section{Introduction}\label{sec:intro}
Deep neural networks (DNNs) have achieved higher performance on various machine-learning tasks. 
However, since computational resources are limited on edge devices such as smartphones, 
it is desirable to optimize the model size suited to a specific device. 
Various methods for compressing DNNs have been proposed, 
including factorized convolutions~\cite{Howard_2017}, low-rank approximations~\cite{Jaderberg_2014}, 
pruning~\cite{Liu_2017}, quantization~\cite{Han_2016}, and neural architecture search~\cite{Tan_2019}.
However, changing the model size is generally difficult for those methods once training and compression are completed; 
even when the same base model is reconfigured on different devices, 
retraining from scratch or fine-tuning according to the resources of the target devices is still required. 
In addition, for multi-task recognition systems operating on devices such as robots, 
it is preferable that the computational load of each task can be dynamically changed according to resource usage.
One option is to store multiple models with different sizes in advance and switch those models, but doing so requires additional storage.
Therefore, it is desirable that the model size can be flexibly changed without retraining, 
which we refer to as {\it scalability} in this paper.
\begin{figure*}[tb]
	\centerline{\includegraphics[width=\linewidth]{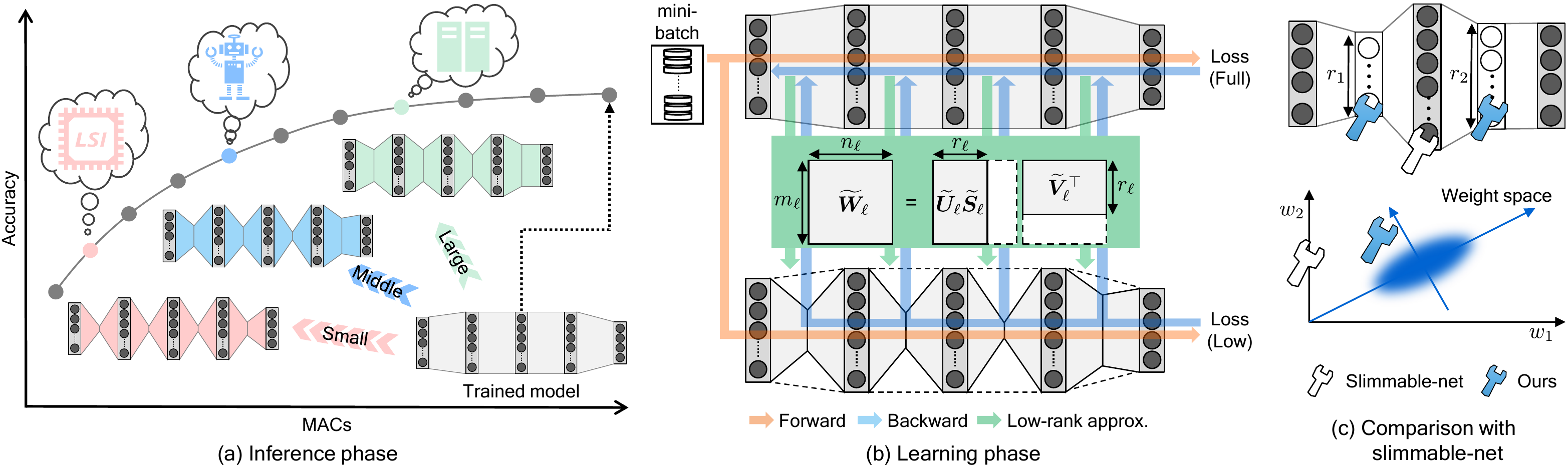}}
	\caption{
		Illustration of Decomposable-Net. 
		These schemes can also be applied to CNNs.
	}
	\label{fig:svd}
\end{figure*}

To that end, \cite{Yu_2019_1} introduced slimmable networks whose width can be changed to predefined ones after training.
Moreover, \cite{Yu_2019_2} proposed universally slimmable networks (US-Nets) that extend slimmable networks to arbitrary widths. 
However, since these methods directly reduce the width (i.e., dimensionality) in each layer, 
the principal components are not taken into account. 
In addition, they uniformly reduce the width across all layers, which ignores differences in the importance of different layers. 

In this paper, we propose Decomposable-Net as a novel scheme that allows DNNs to flexibly change their size after training. 
A weight matrix in each layer is decomposed into two low-rank matrices via singular value decomposition (SVD).
By changing the rank in each layer, Decomposable-Net can scale the model to an arbitrary size (Figure~\ref{fig:svd}(a)).
In contrast to \cite{Yu_2019_2}, we reduce the rank (i.e., the redundant basis in the principal subspace), 
for which the importance of each basis is naturally characterized by a singular value.
This prevents the weight matrix in each layer from losing important directions, as illustrated in Figure~\ref{fig:svd}(c).
Moreover, our approach is essentially different from the existing low-rank compression methods~\cite{Kim_2016,Xu_2019} 
that apply low-rank approximation during or after learning and optimize the model for a specific size. 
Although Decomposable-Net is a single model and is not necessary to retrain, 
we show that it performs significantly better than those methods in a wide range of model sizes. 

Our technical contributions are summarized as follows.
\begin{itemize}
	\item We propose a novel backpropagation scheme that jointly minimizes losses for both of full- and low-rank networks (Figure~\ref{fig:svd}(b)). 
	It is designed not only to maintain the performance of a full-rank network but also to improve multiple low-rank networks (to be used for the inference).
	\item We introduce a simple criterion for the rank selection that utilizes a singular value for each basis, which effectively suppresses approximation error. 
	\item The basic performace with SVD-based (e.g., channel- and spatial-wise) decompositions are enhanced by our methods, in addition to the scalability without retraining.
\end{itemize}

%-------------------------------------------------------------------------------------------------------------------------------------------------------------------------------------
\section{Methods}\label{sec:methods}
\subsection{Overview}
For the $\ell$-th network layer, let $\bm{y}_{\ell} = \bm{W}_{\ell}^\top \bm{x}_{\ell} \in \mathbb{R}^{n_{\ell}}$ be an output vector 
given by the linear transformation of an input vector $\bm{x}_{\ell} = \phi(\bm{y}_{\ell-1}) \in \mathbb{R}^{m_{\ell}}$ 
with the weight matrix $\bm{W}_{\ell} \in \mathbb{R}^{m_{\ell} \times n_{\ell}}$, 
where $\phi(\cdot)$ is the activation function, and $m_{\ell}$ and $n_{\ell}$ are the numbers of input and output nodes, respectively.
Let $R_{\ell} = \min(m_{\ell}, n_{\ell})$ be the full rank of the non-degenerate weight matrix. 
Given $\bm{U}_{\ell} \in \mathbb{R}^{m_{\ell} \times R_{\ell}}$ and $\bm{V}_{\ell} \in \mathbb{R}^{n_{\ell} \times R_{\ell}}$
as matrices that have left and right singular vectors (i.e., bases) as columns, 
and $\bm{S}_{\ell} \in \mathbb{R}^{R_{\ell} \times R_{\ell}}$ as a diagonal matrix composed of singular values~($\sigma_{\ell}$), 
we can formulate the SVD as $\bm{W}_{\ell} = \bm{U}_{\ell} \bm{S}_{\ell} \bm{V}_{\ell}^\top$.

An example of Decomposable-Net with fully connected layers is shown in Figure~\ref{fig:svd}.
Each weight matrix in the network is decomposed into two matrices of rank $R_{\ell}$ via SVD, and we control the rank to change the model size.
This can be viewed as inserting a sub-layer between the original layers and changing its width from $R_{\ell}$.
The number of parameters in each layer becomes $\left(m_{\ell}+n_{\ell}\right)R_{\ell}$ by this decomposition.
Thus, we can compress the network to an arbitrary size by changing the rank $r_{\ell}~(\leq R_{\ell})$ 
within the range $1 \leq r_{\ell} < m_{\ell}n_{\ell} / \left(m_{\ell}+n_{\ell}\right)$.
For CNNs, we transform convolutional tensors to the matrix form and apply the low-rank approximation to every convolutional and fully connected layer.
Note that our methods do not rely on a specific decomposition form, and the SVD-based methods, 
including channel- and spatial-wise~\cite{Zhang_2016,Tai_2016} decompositions, can be adopted.
However, these low-rank approximations solely deteriorate the performance. 
Thus, we propose a learning method to alleviate it. 

\subsection{Learning}\label{subsec:learning}
For the $L$-layer network, let ${\mathcal W} = \{ \bm{W}_{\ell} \}_{\ell=1}^L$ be a set of weight matrices and let 
$\widetilde{\mathcal W} = \{ \widetilde{\bm{W}}_{\ell} \}_{\ell=1}^L$ be a set of their low-rank approximations, respectively.
As shown in Figure~\ref{fig:svd}(b), we additionally propagate each mini-batch to a low-rank network 
whose weight ${\widetilde {\mathcal W}}$ is generated from ${\mathcal W}$ via SVD.
Other trainable parameters $\Theta$~(e.g., biases) are shared between the full- and low-rank networks 
\footnote{
For the batch normalization (BN) layer, $\gamma$ and $\beta$ are shared. 
$\mu$ and $\sigma^2$ are not shared and are not tracked by moving average during training. 
These are computed on the whole training set for each model after training, as in US-Nets.
}.
Therefore, the number of total parameters is the same as in normal learning.

For the purpose to maximize the performance of smaller subnetworks, 
we propose to jointly minimize losses for both full- and low-rank networks as follows: 
\begin{linenomath}
\begin{align}\label{eq:objective}
\min_{{\mathcal W}, \Theta} 
&\frac{1}{B} \sum_{b=1}^B 
 \left\{ 
  (1 - \lambda) \underbrace{{\mathcal L} ( {\mathcal D}_b, {\mathcal W}, \Theta )}_{\text{loss (full-rank)}} 
  + \lambda \underbrace{{\mathcal L} ( {\mathcal D}_b, \widetilde{\mathcal W}, \Theta )}_{\text{loss (low-rank)}}
 \right\} \nonumber \\
&+ \eta {\mathcal R}\left({\mathcal W} \right).
\end{align}
\end{linenomath}
Here, ${\mathcal L}(\cdot)$ is a loss function, ${\mathcal D}_b$ is a training sample in a mini-batch, 
$B$ is the batch size, and $\lambda \in [0, 1]$ is a hyperparameter for balancing between the two losses.
${\mathcal R}(\cdot)$ is a regularization function, 
for which we adopt the sum of squared Frobenius norms ${\mathcal R}({\mathcal W}) = \frac{1}{2}\sum_{\ell=1}^{L} \|\bm{W}_{\ell}\|_F^2$, 
and $\eta$ is a hyperparameter to control its strength.
In order to learn models in an end-to-end manner, 
backpropagation through SVD is involved in computing $\bm{\nabla}_{\ell} := \partial {\mathcal L} ( {\mathcal D}_b, \widetilde{\mathcal W}, \Theta ) / \partial \bm{W}_{\ell}$. 
The detailed derivation of the gradient and a practical technique for improving the numerical stability are given in appendix~\ref{app:a}.
The gradients for $\Theta$ are simply computed from the $\lambda$-weighted average of both networks.

Since we need a single model that performs well at multiple sizes, 
we aim to optimize all the low-rank networks to be selected at the inference phase. 
However, it is computationally intractable to do so in each iteration step. 
Thus, we select one low-rank network with a random size from a predefined range.
Specifically, a ratio $Z$ of the number of bases (referred to as {\it rank ratio}) is sampled from uniform distribution ${\mathcal U} (\alpha_l, \alpha_u)$ with $0 < \alpha_l < \alpha_u \leq 1$. 
Then, we reduce $d=(1-Z)\sum_{\ell=1}^L{R_{\ell}}$ bases across the entire network using a criterion described in the next subsection.
The pseudo-code for learning Decomposable-Net is given in Algorithm~\ref{alg:learning}.

\subsection{Rank Selection}\label{subsec:ranking}
At the stage of compression, 
it is critically important to select an appropriate rank for maximizing the performance (as we see in Figure~\ref{fig:ablation}(c)). 
Given a target size for a model, we select the rank of each layer by reference to a simple criterion: 
$C(\ell, i) = \sigma_{\ell, i}$ (a singular value), where $i \in \{1, \dots, R_{\ell}\}$ is a basis index.
In practice, we globally sort the singular values of all bases in the entire network in ascending order and store them as a single list. 
The only step necessary for selection is to reduce the first $d$ bases in the list. 
We apply the same criterion to learning and inference.
$d$ is determined by the rank ratio $Z$ during learning and the target model size for inference.

The impact of removing an element $w$ of $\bm{W}$ on a loss $\mathcal{L}(\bm{W})$
can be approximated via the first-order Taylor expansion: 
$\mathcal{L}(\bm{W})|_{w=0} - \mathcal{L}(\bm{W}) \approx -w \cdot (\partial\mathcal{L}(\bm{W}) / \partial w)$, 
and its absolute value can be used as an importance measure for pruning~\cite{Lubana_2020}. 
In pruning during learning, \cite{Lubana_2020} showed that keeping large weights results in faster convergence, 
and using only the magnitude $|w|$ works well as the importance. 
According to that, we can characterize the importance of each basis by 
$\sigma_{\ell,i} \cdot \left|\partial\mathcal{L}(\bm{W}_{\ell}) / \partial \sigma_{\ell,i}\right|$, 
which is proportional to the magnitude of singular value $\sigma_{\ell,i}$. 
Therefore, it is expected that our criterion simply using singular values can supress the impact on the loss (i.e., approximation error). 
Note that our criterion is different from the energy-based criterion~\cite{Alvarez_2017,Xu_2019}: 
$C^{\prime}(\ell, i) = \sum_{k=1}^{i} \sigma_{\ell, k}^2 / \sum_{j=1}^{R_{\ell}} \sigma_{\ell, j}^2$. 
Because it evaluates the accumulated energy ratio for each matrix, 
the importance characterized by the magnitude of singular value is ignored. 
In section~\ref{sec:experiments}, we experimentally show that our criterion outperforms it.

\begin{algorithm}[t]
	\caption{Learning Decomposable-Net}\label{alg:learning}
	\begin{algorithmic}[1]
	\REQUIRE Training set $\mathcal{S}$, loss function $\mathcal{L}$, 
	neural network $\mathcal{N}$ with a set of weight matrices ${\mathcal W} = \{ \bm{W}_{\ell} \}_{\ell=1}^{L}$ in which $\bm{W}_{\ell}$ has rank $R_{\ell}$, and other trainable parameters $\Theta$. 
	Regularization parameter $\eta$, 
	balancing parameter $\lambda \in [0, 1]$, 
	lower and upper rank ratios of $\alpha_l$ and $\alpha_u$ ($0 < \alpha_l < \alpha_u \leq 1$), 
	batch size $B$, 
	and criterion $C$ for rank selection
	\ENSURE Return $\mathcal{N}$ with learned $\mathcal{W}$ and $\Theta$
	\STATE{Initialize $\mathcal{W}$ and $\Theta$}
	\REPEAT
	\STATE{Draw $Z$ from the uniform distribution $\mathcal{U}(\alpha_l, \alpha_u)$}
	\STATE{Select the ranks of the weight matrices $\{r_{\ell}\}_{\ell=1}^{L}$ by reducing $d = (1-Z)\sum_{\ell=1}^{L}R_{\ell}$ bases across the entire network, according to the criterion $C$}
	\STATE{Build a low-rank network with a set of weight matrices $\widetilde{\mathcal{W}}$, via rank-$r_{\ell}$ approximation of $\bm{W}_{\ell}$ for $\ell \in \{1, \dots, L\}$}
	\STATE{Draw a batch of samples $\{\mathcal{D}_b\}_{b=1}^{B}$ from $\mathcal{S}$}
	\STATE{Compute the gradients of the objective function in Eq.~(\ref{eq:objective}), with respect to $\bm{W}_{\ell} \in \mathcal{W}$ and $\theta \in \Theta$}
	\STATE{Update $\bm{W}_{\ell} \in \mathcal{W}$ and $\theta \in \Theta$ with the gradients}
	\UNTIL{stopping criterion is met}
	\end{algorithmic}
\end{algorithm}

%-------------------------------------------------------------------------------------------------------------------------------------------------------------------------------------
\subsection{Analysis}\label{subsec:theory}
Here, we describe the mathematical interpretation for our methods and show that Decomposable-Net possesses favorable properties as a scalable model.
Proofs of the propositions described below are given in appendix~\ref{app:b} and \ref{app:c}. 
\begin{proposition}\label{prop:mono}
Let $\widetilde{\bm{W}}_{\ell}(r_{\ell})$ be a rank-$r_{\ell}$ approximation for $\bm{W}_{\ell}$ via SVD, and 
let $\hat{\bm{y}}_{\ell}(r_{\ell}) = \widetilde{\bm{W}}_{\ell}(r_{\ell})^{\top} \bm{x}_{\ell}$. 
Then, the squared error between an original $\bm{y}_{\ell}$ and $\hat{\bm{y}}_{\ell}(r_{\ell})$ satisfies 
$\| \bm{y}_{\ell} - \hat{\bm{y}}_{\ell}(1) \|^2 \geq \dots \geq \| \bm{y}_{\ell} - \hat{\bm{y}}_{\ell}(r_{\ell}) \|^2 \geq 
 \| \bm{y}_{\ell} - \hat{\bm{y}}_{\ell}(r_{\ell}+1) \|^2 \geq \dots \geq \| \bm{y}_{\ell} - \hat{\bm{y}}_{\ell}(R_{\ell}) \|^2 = 0.$
\end{proposition}
\noindent
Therefore, the errors between the original output $\bm{y}_{\ell}$ and 
its approximation $\hat{\bm{y}}_{\ell}$ monotonically decrease as the rank increases. 
Although this is a natural property of SVD, it is desirable for changing the model size with maintaining the performance. 
It does not exactly hold for the entire network, 
but we can show that changes of the upper bound depends on the layer-wise error $\| \bm{y}_{\ell} - \hat{\bm{y}}_{\ell} \|^2$:
\begin{proposition}\label{prop:bound_l2}
For a fully connected network with 1-Lipschitz activation function (e.g., ReLU), 
let $\bm{p}$ and $\widetilde{\bm{p}}$
be $K$-class softmax outputs from the full- and low-rank networks, respectively.
Then, KL divergence between them can be bounded as ${\rm KL}(\bm{p}||\widetilde{\bm{p}}) \leq$
\small
\begin{linenomath}
\begin{align*}
\sqrt{2 \left( \|\bm{y}_L - \hat{\bm{y}}_L\|^2 + \sum_{\ell=1}^{L-1} \left\{ \|\bm{y}_{\ell}-\hat{\bm{y}}_{\ell}\|^2 \prod_{j=\ell+1}^{L} \|\bm{W}_{j}\|_2^2 \right\} \right)}, 
\end{align*}
\end{linenomath}
\normalsize
where $\|\cdot\|_2$ indicates the spectral norm.
\end{proposition}
\noindent
Hence, it can be expected that the performance of entire network changes in conjunction with the rank in each layer. 
Regarding the Lipschitz constant $\prod_{j=\ell+1}^{L} \|\bm{W}_{j}\|_2^2$, 
we experimentally verified that the empirical value is much smaller than the theoretical one~\cite{Wei_2019} (see appendix~\ref{app:e}).

Under the condition that a trained network has near low-rank weight matrices, 
\cite{Arora_2018,Suzuki_2020,Suzuki_2019} proved that the condition contributes not only to compressing the network efficiently, 
but also to yielding a better generalization error bound for the non-compressed (full-rank) network. 
That motivates our approach, namely, 
a learning scheme that aims to facilitate the performance of low-rank networks as well as that of a full-rank network. 

%-------------------------------------------------------------------------------------------------------------------------------------------------------------------------------------
\section{Related Work}\label{sec:related_work}
\subsection{Low-Rank Approximations}
Compression methods based on low-rank approximation have been proposed in the literature, 
and those can be divided into three categories.
(1) Before learning, 
\cite{Ioannou_2016} factorized a convolutional kernel into horizontal and vertical kernels.
However, the rank in each layer needs to be manually selected.
(2) After learning, 
\cite{Jaderberg_2014,Zhang_2016} applied data-dependent optimization with the low-rank constraint. 
Alternatively, tensor methods (e.g., CP and Tucker decompositions) that directly approximate four-way convolutional tensors 
have been adopted~\cite{Lebedev_2015,Kim_2016}.
However, since the approximation is applied after learning the model in each of these methods, 
the low-rankness is not necessarily ensured.
Moreover, they optimize networks under a given target size, 
and thus optionally require retraining or fine-tuning to reconfigure the models for different devices. 
(3) During learning, 
\cite{Alvarez_2017,Xu_2019} utilized trace-norm regularization as a low-rank constraint. 
The performance of these methods depends on a hyperparameter for adjusting the strength of regularization.
It is difficult to decide on an appropriate range for the hyperparameter in advance, 
meaning that selection requires trial and error to achieve a particular model size.

Our methods are related to (3), in particular to Trained Rank Pruning (TRP)~\cite{Xu_2019}. 
TRP minimizes the loss only for the low-rank network produced by SVD.
Since it applies trace-norm regularization and truncates small singular values during learning, 
the resulting ranks in the low-rank network are fixed.
Thus, the network does not perform well with other ranks, as shown in section~\ref{sec:experiments}.

\subsection{Scalable Networks}
\cite{Yu_2019_1,Yu_2019_2} proposed slimmable networks and US-Nets, which are scalable in the width direction.
These works are closely related to ours, but there are differences in some aspects.
First, since these methods directly and uniformly reduce the width for every layer, 
the principal components are not taken into account, and the relative importance of each layer is not considered.
Second, although the US-Nets allow arbitrary changes in width, 
monotonicity of error is not guaranteed when changing widths in each layer.

Recent works further acquired scalability in depth~\cite{Zhang_2019}, resolution~\cite{Cai_2020,Yu_2020,Yang_2020}, and kernel size~\cite{Cai_2020,Yu_2020}.
The proposed methods are focused on improving the width scalability, but can be complementarily combined with those directions. 
This topic remains as our future works.

%-------------------------------------------------------------------------------------------------------------------------------------------------------------------------------------
\section{Experiments}\label{sec:experiments}
\begin{figure*}[t]
	\centerline{\includegraphics[width=\linewidth]{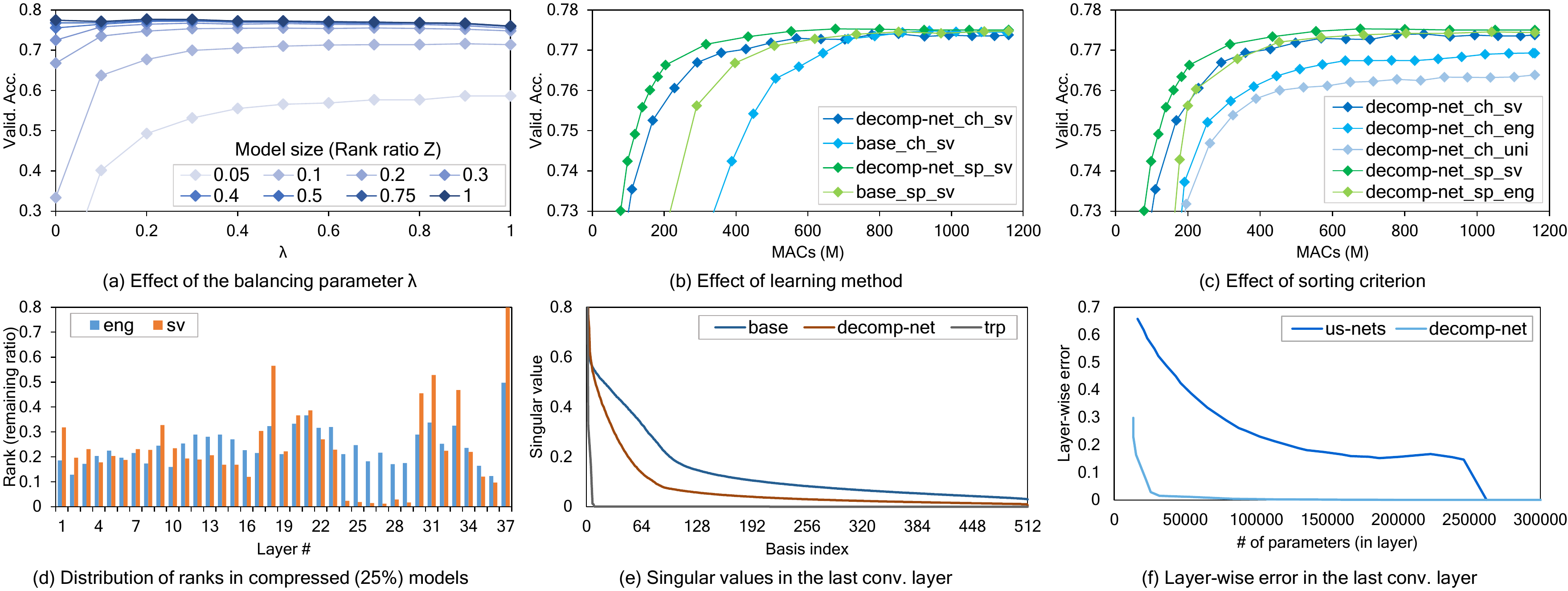}}
	\caption{
		Ablation study with ResNet-34 on CIFAR-100. 
		\texttt{base} indicates normal learning as our baseline, likewise \texttt{decomp-net} indicates the proposed method.
		\texttt{ch} and \texttt{sp} indicate channel- and spatial-wise decompositions, and 
		\texttt{uni}, \texttt{eng} and \texttt{sv} indicate uniform rank selection, energy-based and our singular value-based criterion, respectively.
	}
	\label{fig:ablation}
\end{figure*}
\begin{figure*}[t]
	\centerline{\includegraphics[width=\linewidth]{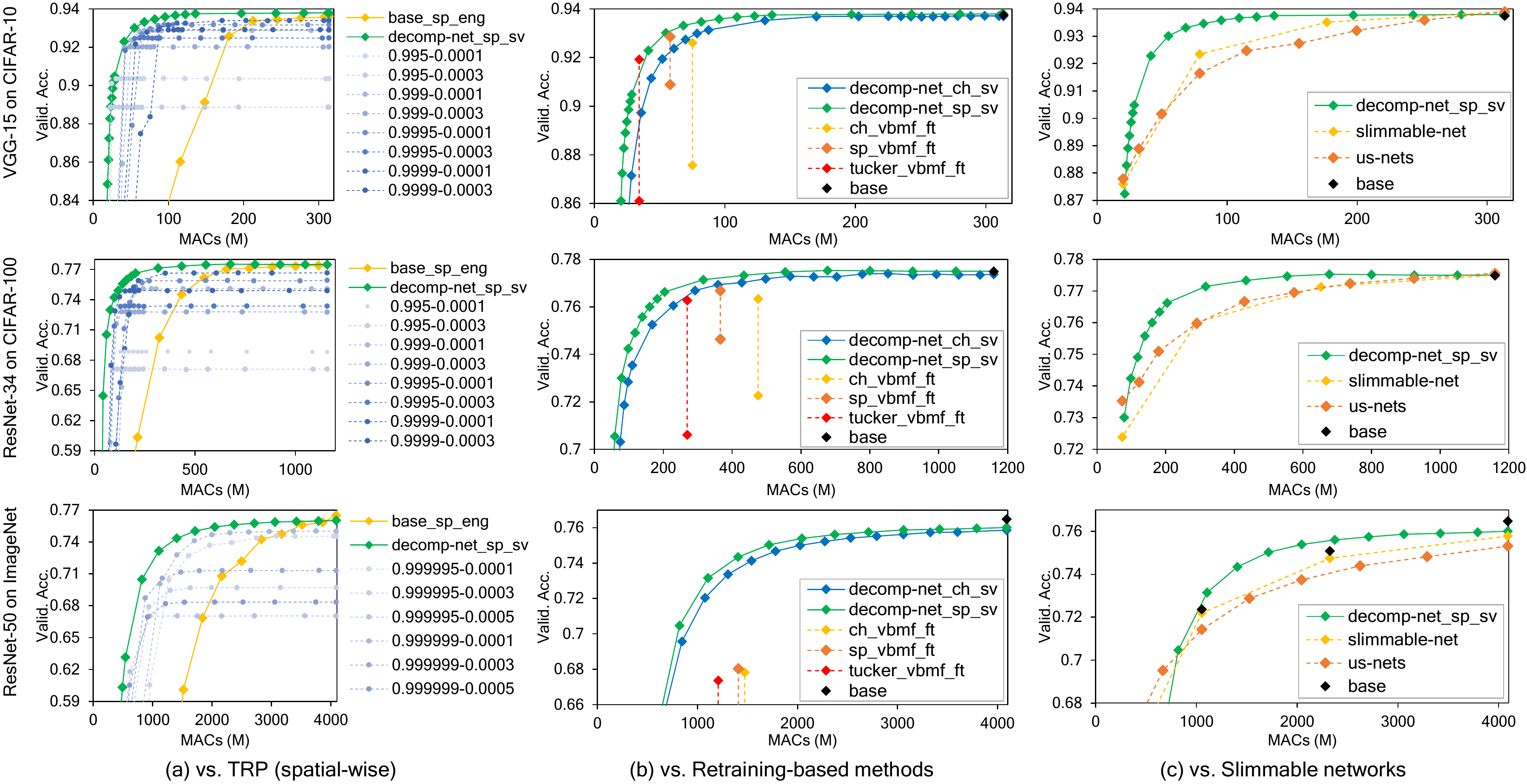}}
	\caption{
		Comparison results: (top) VGG-15 on CIFAR-10, (middle) ResNet-34 on CIFAR-100, and (bottom) ResNet-50 on ImageNet. 
		(a) shows the results vs. TRP~\protect\cite{Xu_2019} with spatial-wise decomposition. 
		Dashed lines are the results from TRP, individually trained with different parameters ($e-\tau$).
		\texttt{base\_sp\_eng} is a baselline, in which normal learning is adopted instead of TRP.
		(b) shows the results vs. retraining-based methods.
		\texttt{base} is a baseline for the full model.
		\texttt{tucker\_vbmf\_ft} indicates the method by \protect\cite{Kim_2016}.
		Dashed vertical lines show the performance improvements of retraining.
		(c) shows the results vs. slimmable networks~\protect\cite{Yu_2019_1} and US-Nets~\protect\cite{Yu_2019_2}.
		For the ImageNet dataset, three baseline models are trained with diffrent ratios of the number of channels ($1.0\times$, $0.75\times$, and $0.5\times$).
	}
	\label{fig:comparison}
\end{figure*}

We evaluate our methods on image-classification tasks of CIFAR-10/100~\cite{Krizhevsky_2009} and ImageNet~\cite{Deng_2009} datasets using deep CNNs.
For the CIFAR-10 dataset, we adopt a VGG-like network with fifteen layers (VGG-15)~\cite{Zagoruyko_2015,Liu_2017}. 
For the CIFAR-100 and ImageNet datasets, we adopt ResNet-34 and ResNet-50~\cite{He_2016}, respectively. 
We follow the same baseline setup for the CIFAR datasets as used by \cite{Zagoruyko_2016}, 
and the setup of \cite{Yu_2019_1} for the ImageNet dataset. 
The additional details of experiments are described in appendix~\ref{app:d}.

For all experiments, we report the results of the models after the last epoch. 
The results are averaged over multiple runs with different random seeds, specifically, with five seeds $(0 \sim 4)$ on the CIFAR datasets and three seeds $(0 \sim 2)$ on the ImageNet dataset.
All methods are evaluated in terms of the tradeoff between validation (top-1) accuracy and the number of multiply-accumulate operations (MACs). 
The number of MACs is defined as $\sum_{\ell=1}^L P_{\ell}H_{\ell}W_{\ell}$, where $P_{\ell}$, $H_{\ell}$, and $W_{\ell}$ are the number of parameters, and the height and width of an output feature map, respectively.
We compare channel- and spatial-wise decompositions for convolutional layers, 
but use the channel-wise one for $1 \times 1$ kernels, since they have no spatial redundancy. 
We experimentally tuned hyperparameters, and set $\alpha_u=0.25$, $0.5$, and $0.8$, respectively for CIFAR-10, CIFAR-100, and ImageNet datasets, while fixing $\alpha_l=0.01$ for all datasets.
Except for the results shown in Figure~\ref{fig:ablation}(a), we set $\lambda=0.5$ to balance the performance of a full- and low-rank models. 
We implemented other methods in the following experiments and evaluated them under our settings. 

\subsection{Ablation Study}\label{subsec:ablation}
We verify each component in our methods with ResNet-34 on the CIFAR-100 dataset.
Firstly, we evaluate our learning method with various values for parameters $\lambda \in [0, 1]$ that balance between two losses. 
Each model learned with a specific $\lambda$ is evaluated for several sizes of the rank ratio $Z \in \{0.05, 0.1, 0.2, 0.3, 0.4, 0.5, 0.75, 1\}$ without retraining.
The results are illustrated in Figure~\ref{fig:ablation}(a).
Each curve in the figure shows the performance changes in a specific model size (e.g., \texttt{0.05} and \texttt{1} correspond to $5\%$ and the full model, respectively) 
against $\lambda$, where points at $\lambda=0$ correspond to results from normal learning.
We can see that the performances of compressed models improve as $\lambda$ increases, while that of the full model is almost flat.
This implies that our learning method not only maintains the performance of full-rank network, but also improves multiple low-rank networks in a single model.
Notably, the model with $\lambda=0.3$ achieves $77.65\%$ at full size, which is better than the $77.49\%$ under normal learning, 
demonstrating that the compressible network generalizes well~\cite{Suzuki_2019}. 
The effect of our method is also verified in Figure~\ref{fig:ablation}(b), 
where ours (\texttt{decomp-net}) is consistently better than normal learning (\texttt{base}), with both channel- (\texttt{ch}) and spatial-wise (\texttt{sp}) decompositions. 

Secondly, we evaluate the criterion for rank selection with fixed learning method.
A method that uniformly reduce the rank across all layers (\texttt{uni}) and the energy-based method~\cite{Alvarez_2017,Xu_2019} (\texttt{eng}) 
are compared with ours (\texttt{sv}) in Figure~\ref{fig:ablation}(c). 
While \texttt{eng} is better than \texttt{uni}, 
\texttt{sv} successfully maintains the performance and is the best among them. 
The difference in selected ranks is illustrated in Figure~\ref{fig:ablation}(d).
It can be observed that the rank distribution of \texttt{sv} is sharper than that of \texttt{eng}, which is near-uniform.
We consider that \texttt{eng} does not fully characterize the importance of each basis, 
since it ignores the differences of the magnitude of singular values accross layers.

\subsection{Comparison with Low-Rank Networks}\label{subsec:comparison1}
First, we compare Decomposable-Net with TRP~\cite{Xu_2019}. 
We implemented TRP referencing to their code\footnote{\texttt{\url{https://github.com/yuhuixu1993/Trained-Rank-Pruning}}}.
TRP has three parameters, namely, the strength $\tau$ of trace norm regularization, a threshold $e$ for truncating singular values, and a truncation interval $T$.
TRP optimizes a model of a specific size, but the size of an obtained model depends on those parameters.
Thus, we compare several models individually trained with $\tau \in \{0.0001, 0.0003\}$ and $e \in \{0.995, 0.999, 0.9995, 0.9999\}$ for the CIFAR datasets, 
and $\tau \in \{0.0001, 0.0003, 0.0005\}$ and $e \in \{0.999995, 0.999999\}$ for the ImageNet dataset.
We report the results with $T=1$ since it was the best among $\{1, 2, 4, 8\}$ for all datasets. 

The results with spatial-wise decomposition are summarized in Figure~\ref{fig:comparison}(a) 
(the additional results with channel-wise decomposition are illustrated in Figure~\ref{fig:comparison_sm1} in the appendix). 
TRP adopts the energy-based criterion for rank selection, 
and thus we compress their models with it at several sizes after training. 
Since TRP truncates small singular values during learning, the resulting ranks in the low-rank network are fixed.
Thus, the network does not perform well at sizes other than obtained ranks, as shown in the figure.
In contrast, although Decomposable-Net is a single model, it is basically better at almost all sizes. 
Moreover, our methods do not require exhaustive parameter searches to yield a good model at a specific size, 
since they require training only one time and changing size.
Figure~\ref{fig:ablation}(e) shows an example of singular values obtained from each learning method.
The singular values with TRP rapidly decrease to zero, because TRP applies trace-norm regularization.
The bases with a singular value of zero can be reduced without inducing errors.
However, we consider that this limits the representational power of larger models, 
as we can see the flat lines for TRP in Figure~\ref{fig:comparison}(a).
On the other hand, our methods reduce singular values in comparison with normal learning, but do not push them to zero.
We see that this contributes to improving the performance of low-rank networks while maintaining that of the full-rank network.

Next, we compare with retraining-based methods in which low-rank approximation and fine-tuning are applied after normal learning.
Specifically, we adopt a method by \cite{Kim_2016} that directly approximates convolutional tensors with Tucker decomposition, 
where variational Bayesian matrix factorization (VBMF)~\cite{Nakajima_2012} is applied to determine the ranks.
We used publicly available code\footnote{\texttt{\url{https://sites.google.com/site/shinnkj23/downloads}}} for VBMF.
For Tucker decomposition, we apply higher-order orthogonal iteration~\cite{Lathauwer_2000} under given ranks. 
Then, we perform fine-tuning with a grid search for selecting training parameters. 
The ranges for the search and the obtained parameters are described in appendix~\ref{app:d}.

The results with the best parameters are shown in Figure~\ref{fig:comparison}(b).
We additionally evaluate channel- (\texttt{ch}) and spatial-wise (\texttt{sp}) decompositions by switching them and individually optimizing the parameters. 
Note that \texttt{sp\_vbmf\_ft} corresponds to the method by \cite{Tai_2016}, in which the rank selection method is changed from manual to VBMF.
It can be observed that the performance significantly drops from \texttt{base} after decomposition, 
but can be recovered by retraining (dashed vertical lines). 
This is because the low-rankness is not ensured in the models trained with normal learning. 
Tucker decomposition is more efficient than other ones and is comparable to ours at a specific size on the CIFAR datasets. 
However, the performance drops when the size is changed, 
and retraining is required to recover the performance again.
For practical use, such methods are not suited to reconfigure the same model on other devices with different resources.
In this paper, we adopt the matrix (channel- and spatial-wise) decompositions, 
but it is possible to extend our methods with Tucker decomposition since it is a generalization of SVD to higher orders.

\subsection{Comparison with Slimmable Networks}\label{subsec:comparison2}
We compare Decomposable-Net with slimmable networks~\cite{Yu_2019_1} and US-Nets~\cite{Yu_2019_2}. 
We implemented them by referencing their code\footnote{\texttt{\url{https://github.com/JiahuiYu/slimmable_networks}}}. 
Slimmable networks are trained at four fixed widths of ratios $\{0.25, 0.5, 0.75, 1\}$.
US-Nets are trained with a lower width of ratio 0.25, upper width of ratio 1, and two random widths between them. 
In-place distillation~\cite{Yu_2019_2} is applied to US-Nets.
The original loss of slimmable networks is an un-weighted sum of all training losses with different widths. 
Although it works well with ResNet-50 on the ImageNet dataset, 
the performance on the CIFAR datasets is unsatisfactory especially when comparing the accuracy (76.6\%) of a full-width ResNet-34 with the baseline (77.5\%) on the CIFAR-100 dataset. 
Therefore, we use the average instead of the sum on the CIFAR datasets.

The results are shown in Figure~\ref{fig:comparison}(c).
For the CIFAR datasets, all methods have performance with the full model comparable to that of the baseline (\texttt{base}). 
However, we can see that the performances of slimmable networks and US-Nets decrease as MACs reduce, which are slightly faster than ours.
This can be interpreted as those methods directly reducing the width (i.e., dimensionality) in each layer, and thus losing important directions in the weight space.
To better understand this property, we illustrate an example of layer-wise error in Figure~\ref{fig:ablation}(f).
It can be seen that the error for US-Nets rapidly increases as the number of parameters decreases in the layer, and this increase is not monotonic.
In comparison, the error for Decomposable-Net increases slower than that for US-Nets, and monotonicity is ensured as expected.
We see this is due to taking the principal components into account for Decomposable-Net, which contributes to maintaining performance over a wide range of model sizes.
Moreover, we believe our criterion, which ununiformly select ranks in the network (as we see in Figure~\ref{fig:ablation}(d)) contributes too, 
in contrast to slimmable networks and US-Nets, which reduce the width uniformly across all layers.

For the ImageNet dataset, we can see that the accuracy of Decomposable-Net is maintained for the full model better than under slimmable networks and US-Nets. 
Specifically, the accuracy of the full model is $76.48\%$ for the baseline and $76.01\%$ for Decomposable-Net, which is better than $75.31\%$ for US-Nets. 
Moreover, Decomposable-Net achieves better accuracy than the baselines individually trained with $0.75\times$ and $0.5\times$ number of channels, 
while slimmable networks and US-Nets show lower accuracies than the baselines. 
It can be also observed that Decomposable-Net is better up to the model with around $0.2\times$MACs, 
but lower than slimmable networks and US-Nets when the compression rate is extremely high.
Our methods use SVD and reduce the bases, which means it does not change the number of inputs and outputs (i.e., the input and output dimensionalities).
The number of parameters in each layer is $(m_{\ell}+n_{\ell})r_{\ell}$, thus decreasing linearly with respect to the rank.
US-Nets reduces both input and output dimensionality, meaning the number of parameters decreases at a quadratic rate.
This makes it easier for US-Nets to achieve extremely high compression.
Therefore, it can be worthwhile to combine our methods with pruning to reduce the number of input or output channels, 
by incorporating a channel-level sparseness constraint during learning~\cite{Alvarez_2017}. 
We will consider this for future improvements.

\subsection{Remarks}\label{subsec:remarks}
The results on the ImageNet dataset are summarized in Table~\ref{tbl:summary}, 
where we picked up results from the methods compared in this paper, for around $0.25\times$MACs models.
To further validate our learning method, 
we decomposed ResNet-50 with the same ranks (i.e., exactly same MACs) as Decomposable-Net in Table~\ref{tbl:summary}. 
Then, we trained it normally from scratch, and obtained the top-1 accuracy of $71.55 \pm 0.17\%$.
Therefore, we see that the performance of our method ($73.15 \pm 0.06\%$) is not easily achieved with the normal training, 
even by specializing the model to a specific size.

The wall-clock time for training Decomposable-Net with ResNet-50 on the ImageNet dataset was 6.9 days with eight NVIDIA V100 GPUs, 
whereas TRP (that also applies SVD during training) and US-Nets took 5.7 and 6.9 days, respectively. 
Therefore, the training cost of Decomposable-Net is not particularly high and comparable to them.
Additionally, we measured the actual inference time per image with Intel Core i7-8700K CPU on the CIFAR-100 dataset. 
We used approximately $0.25\times$MACs model for Decomposable-Net with ResNet-34.
For the other methods, we chose the models closest to ours in terms of MACs or accuracy. 
As shown in Table~\ref{tbl:time}, 
we can see that Decomposable-Net is as fast as Tucker decomposition with the accuracy comparable to US-Nets.

\begin{table}[tb]
\begin{center}
	\small
	\begin{tabularx}{\linewidth}{lCC}
	\bhline{1pt}
	Method							&	Acc.~(\%)					&	MACs~($\times$)		\\
	\hline
	Baseline~($1.0\times$)			&	$76.48 \pm 0.09$			&	$1.00 \pm 0.000$	\\
	Baseline~($0.5\times$)			&	$72.37 \pm 0.11$			&	$0.26 \pm 0.000$	\\
	TRP~(\texttt{sp})~\cite{Xu_2019}&	$70.57 \pm 0.22$			&	$0.28 \pm 0.005$	\\
	Tucker~\cite{Kim_2016}			&	$67.36 \pm 0.06$			&	$0.30 \pm 0.005$	\\
	Slimmable-net~\cite{Yu_2019_1}	&	$72.19 \pm 0.12$			&	$0.26 \pm 0.000$	\\
	US-Nets~\cite{Yu_2019_2}		&	$71.43 \pm 0.12$			&	$0.26 \pm 0.000$	\\
	Decomposable-Net~(\texttt{sp})	&	$\textbf{73.15} \pm 0.06$	&	$0.27 \pm 0.003$	\\	
	\bhline{1pt}
	\end{tabularx}
	\caption{Results with ResNet-50 on the ImageNet dataset.}
	\label{tbl:summary}
\end{center}
\end{table}
\begin{table}[tb]
\begin{center}
	\small
	\begin{tabularx}{\linewidth}{lCCC}
	\bhline{1pt}
	Method							&	Acc.~(\%)	&	Time~(ms)	&	Speed~($\times$)	\\
	\hline
	Baseline						&	77.49		&	4.57		&	1.00				\\
	TRP~(\texttt{sp})~\cite{Xu_2019}&	76.65		&	2.67		&	1.71				\\
	Tucker~\cite{Kim_2016}			&	76.28		&	1.96		&	2.33				\\
	US-Nets~\cite{Yu_2019_2}		&	77.24		&	3.27		&	1.40				\\
	Decomposable-Net~(\texttt{sp})	&	77.15		&	2.02		&	2.26				\\
	\bhline{1pt}
	\end{tabularx}
	\caption{CPU inference time per image with ResNet-34 on the CIFAR-100 dataset.}
	\label{tbl:time}
\end{center}
\end{table}
%-------------------------------------------------------------------------------------------------------------------------------------------------------------------------------------
\section{Conclusions}\label{sec:conclusions}
We proposed a novel method that allows DNNs to flexibly change their size after training, 
by means of the rank of low-rank matrices in each layer.
To simultaneously maintain the performance of a full-rank network and improve multiple low-rank networks in a single model, 
we proposed a novel learning method that explicitly minimizes losses for both of full- and low-rank networks in a differentiable way.
In experiments on multiple image-classification tasks using deep CNNs, 
Decomposable-Net exhibited favorable performance at several sizes of compressed models, 
relative to that of the low-rank compression methods and slimmable networks.

%-------------------------------------------------------------------------------------------------------------------------------------------------------------------------------------
\section*{Acknowledgements}
TS was partially supported by JSPS KAKENHI (18K19793, 18H03201, and 20H00576), and JST CREST.

\bibliographystyle{named}
\bibliography{ijcai21}

%-------------------------------------------------------------------------------------------------------------------------------------------------------------------------------------
\part*{Appendix}
\appendix
\section{Derivation of the Gradient and Practical Implementation}\label{app:a}
Here, we derive the gradient 
$\bm{\nabla}_{\ell} = \partial {\mathcal L} ( {\mathcal D}_b, \widetilde{\mathcal W}, \Theta ) / \partial \bm{W}_{\ell}$ 
in the case where $m_{\ell} \geq n_{\ell}$~(i.e., $R_{\ell}=n_{\ell}$), but this can be applied to matrices with $m_{\ell} < n_{\ell}$ by transposing them.
We omit the subscript $\ell$ on layer indexes for simplicity of notation.

Let $\bm{I}_r$ be the $r \times r$ identity matrix, and let $\bm{O}_{a, b}$ be the $a \times b$ zero matrix. 
Let $\bm{M}_r=\left(\begin{array}{rr}
             \bm{I}_r			&	\bm{O}_{r, n-r} \\
             \bm{O}_{n-r, r}	&	\bm{O}_{n-r, n-r} \\
             \end{array}\right)$.
Then, the truncated-SVD of $\bm{W}$ can be formulated as $\widetilde{\bm{W}}=\bm{U} \bm{S} \bm{M}_r \bm{V}^{\top}$, 
and the following gradients can be derived through matrix calculus.
\begin{linenomath}
\begin{align}
\frac{\partial {\mathcal L} ( {\mathcal D}_b, \widetilde{\mathcal W}, \Theta )}{\partial \bm{U}} &= \widetilde{\bm{\nabla}} \bm{V} \bm{M}_r \bm{S} \\
\frac{\partial {\mathcal L} ( {\mathcal D}_b, \widetilde{\mathcal W}, \Theta )}{\partial \bm{S}} &= \left( \bm{U}^{\top} \widetilde{\bm{\nabla}} \bm{V} \bm{M}_r \right)_{\mathrm{diag}} \\
\frac{\partial {\mathcal L} ( {\mathcal D}_b, \widetilde{\mathcal W}, \Theta )}{\partial \bm{V}} &= \widetilde{\bm{\nabla}}^{\top} \bm{U} \bm{S} \bm{M}_r
\end{align}
\end{linenomath}
$(\cdot)_{\mathrm{diag}}$ is an operator that sets all off-diagonal elements to 0.
$\widetilde{\bm{\nabla}} = \partial {\mathcal L} ( {\mathcal D}_b, \widetilde{\mathcal W}, \Theta ) / \partial \widetilde{\bm{W}}$ 
can be computed by backpropagating loss through the low-rank network up to $\widetilde{\bm{W}}$.
Let $\widetilde{\bm{S}}$ be a matrix having the top $r$ singular values in its diagonal, 
and let $\widetilde{\bm{U}}$ and $\widetilde{\bm{V}}$ be matrices composed of corresponding left and right singular vectors in those columns.
Then, 
$\bm{S}=\left(\begin{array}{rr}
             \widetilde{\bm{S}}	&	\bm{O}_{r, n-r} \\
             \bm{O}_{n-r, r}	&	\overline{\bm{S}} \\
             \end{array}\right)$, 
$\bm{U}=(\widetilde{\bm{U}}, \overline{\bm{U}})$, and $\bm{V}=(\widetilde{\bm{V}}, \overline{\bm{V}})$.
Substituting the above into Eqs.~(16) and (17) in \cite{Ionescu_2015_2}, we have 
\begin{linenomath}
\begin{align}
\bm{\nabla}
&= \widetilde{\bm{\nabla}}
   \widetilde{\bm{V}} \widetilde{\bm{V}}^{\top} 
     + \bm{U} \bm{S} 
       \left(\begin{array}{rr}
             \bm{O}_{r, r}		&	\bm{\Phi} \\
             \bm{\Phi}^{\top}	&	\bm{O}_{n-r, n-r} \\
       \end{array}\right)
       \bm{V}^{\top} \label{eq:a}, \\
\bm{\Phi}
&= \bm{K} \odot \left( \widetilde{\bm{V}}^{\top} \widetilde{\bm{\nabla}}^{\top} \overline{\bm{U}} \overline{\bm{S}}
                     + \widetilde{\bm{S}} \widetilde{\bm{U}}^{\top} \widetilde{\bm{\nabla}} \overline{\bm{V}} \right), \\
(\bm{K})_{ij}
&= \frac{1}{\sigma_{i}^2 - \sigma_{k}^2} \nonumber \\
&(1 \leq i \leq r, 1 \leq j \leq n - r, k = j + r) \label{eq:b}, 
\end{align}
\end{linenomath}
where $\odot$ represents the Hadamard product.

\begin{figure}[t]
	\centerline{\includegraphics[width=\linewidth]{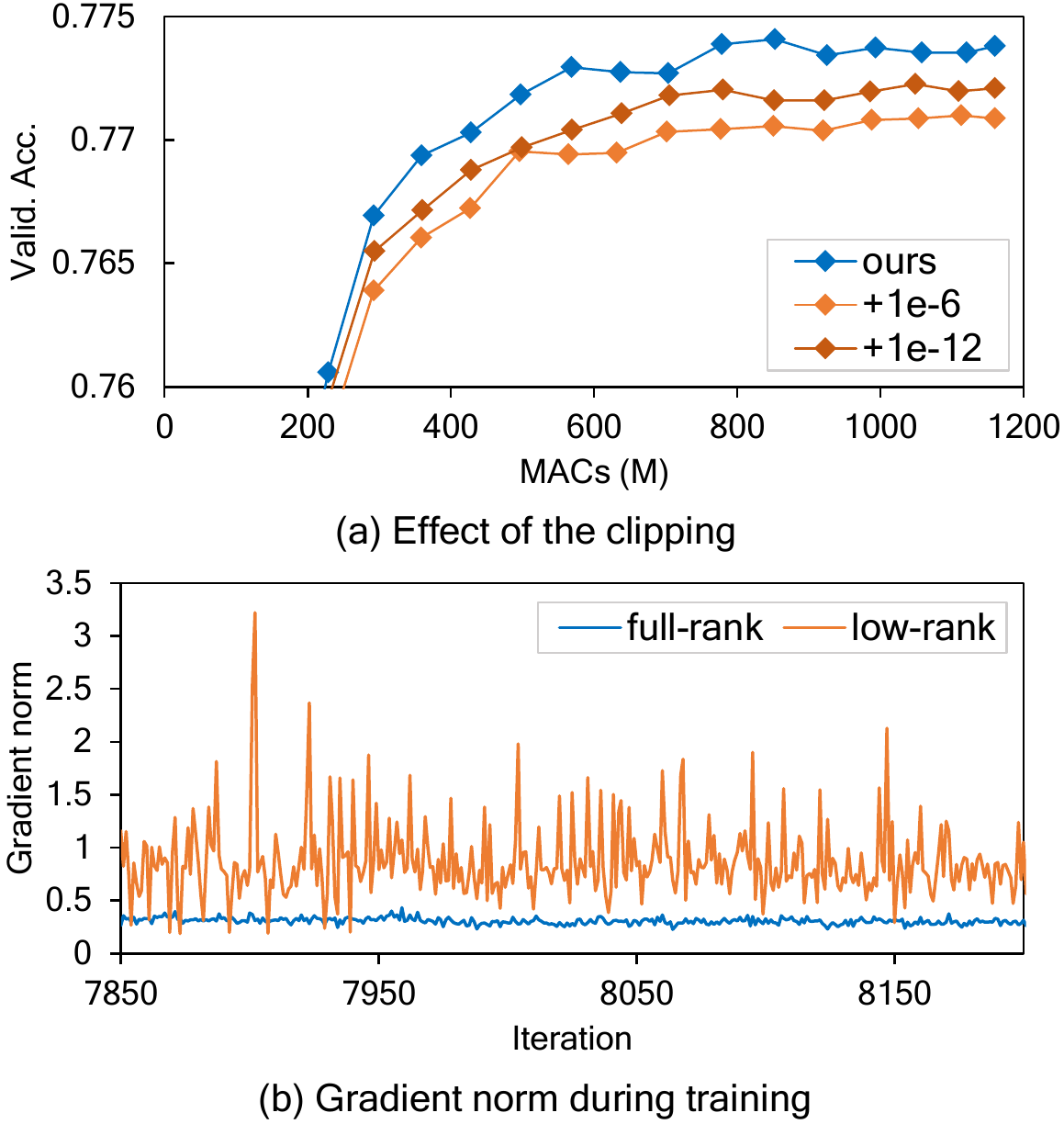}}
	\caption{
		(a) Effect of the clipping with ResNet-34 on CIFAR-100. 
		\texttt{+1.0e-6} and \texttt{+1.0e-12} are results from the method by \protect\cite{Liao_2019}, using small constants $1.0\times10^{-6}$ and $1.0\times10^{-12}$, respectively. 
		(b) An example of gradient norms for full- and low-rank networks.
	}
	\label{fig:eps_norm}
\end{figure}

If $\bm{W}$ does not have the same singular values, $\sigma_{i}^2 > \sigma_{k}^2$ holds in Eq.~(\ref{eq:b}).
However, gaps between neighboring singular values are likely to be small, 
and therefore $(\bm{K})_{ij}$ may diverge in practice. 
Adding or clipping by a small positive constant in the denominator~\cite{Liao_2019,Seeger_2017} has been applied to mitigate this problem, 
but the appropriate value for such a constant depends on the scale of $\bm{W}$ and possibly varies layer by layer in DNNs.
The following shows that the gap in $(\bm{K})_{ij}$ can be represented in scale-invariant form.
Letting $\bm{A}=\widetilde{\bm{V}}^{\top} \widetilde{\bm{\nabla}}^{\top} \overline{\bm{U}}$ and $\bm{B}=\widetilde{\bm{U}}^{\top} \widetilde{\bm{\nabla}} \overline{\bm{V}}$, 
we can rewrite $\bm{\nabla}$ in Eq.~(\ref{eq:a}) as 
\begin{linenomath}
\begin{align}
\bm{\nabla}
&= \widetilde{\bm{\nabla}} \widetilde{\bm{V}} \widetilde{\bm{V}}^{\top} 
   + \widetilde{\bm{U}} \widetilde{\bm{S}} \bm{\Phi} \overline{\bm{V}}^{\top}
   + \overline{\bm{U}} \overline{\bm{S}} \bm{\Phi}^{\top} \widetilde{\bm{V}}^{\top} \nonumber \\
&= \widetilde{\bm{\nabla}} \widetilde{\bm{V}} \widetilde{\bm{V}}^{\top} 
   + \widetilde{\bm{U}} \widetilde{\bm{S}} (\bm{K} \odot \bm{A} \overline{\bm{S}} + \bm{K} \odot \widetilde{\bm{S}} \bm{B}) \overline{\bm{V}}^{\top} \nonumber \\
&\qquad+ \overline{\bm{U}} \overline{\bm{S}} (\bm{K}^{\top} \odot \overline{\bm{S}} \bm{A}^{\top} + \bm{K}^{\top} \odot \bm{B}^{\top} \widetilde{\bm{S}} ) \widetilde{\bm{V}}^{\top} \nonumber \\
&= \widetilde{\bm{\nabla}} \widetilde{\bm{V}} \widetilde{\bm{V}}^{\top} 
   + \widetilde{\bm{U}} (\widetilde{\bm{S}} \bm{K} \overline{\bm{S}} \odot \bm{A} + \widetilde{\bm{S}} \widetilde{\bm{S}} \bm{K} \odot \bm{B}) \overline{\bm{V}}^{\top} \nonumber \\
&\qquad+ \overline{\bm{U}} (\overline{\bm{S}} \overline{\bm{S}} \bm{K}^{\top} \odot \bm{A}^{\top} + \overline{\bm{S}} \bm{K}^{\top} \widetilde{\bm{S}} \odot \bm{B}^{\top} ) \widetilde{\bm{V}}^{\top} \nonumber \\
&= \widetilde{\bm{\nabla}} \widetilde{\bm{V}} \widetilde{\bm{V}}^{\top} 
   + \widetilde{\bm{U}} (\widetilde{\bm{S}} \bm{K} \overline{\bm{S}} \odot \bm{A} + \widetilde{\bm{S}} \widetilde{\bm{S}} \bm{K} \odot \bm{B}) \overline{\bm{V}}^{\top} \nonumber \\
&\qquad\qquad+ \overline{\bm{U}} (\bm{K} \overline{\bm{S}} \overline{\bm{S}} \odot \bm{A} + \widetilde{\bm{S}} \bm{K} \overline{\bm{S}} \odot \bm{B} )^{\top} \widetilde{\bm{V}}^{\top}, 
\end{align}
\end{linenomath}
where 
\begin{linenomath}
\begin{align}
\left( \widetilde{\bm{S}} \bm{K} \overline{\bm{S}} \right)_{ij}		&= \frac{\sigma_i \sigma_k}{\sigma_i^2 - \sigma_k^2}	= \frac{\sigma_k / \sigma_i}{1 - (\sigma_k / \sigma_i)^2}		= \frac{\rho_{ij}}{1 - \rho_{ij}^2}, \\
\left( \widetilde{\bm{S}} \widetilde{\bm{S}} \bm{K} \right)_{ij}	&= \frac{\sigma_i^2}{\sigma_i^2 - \sigma_k^2}			= \frac{1}{1 - (\sigma_k / \sigma_i)^2}							= \frac{1}{1 - \rho_{ij}^2}, \\
\left( \bm{K} \overline{\bm{S}} \overline{\bm{S}} \right)_{ij}		&= \frac{\sigma_k^2}{\sigma_i^2 - \sigma_k^2}			= \frac{(\sigma_k / \sigma_i)^2}{1 - (\sigma_k / \sigma_i)^2}	= \frac{\rho_{ij}^2}{1 - \rho_{ij}^2}.
\end{align}
\end{linenomath}
Hence, the above equations are characterized by the ratio of singular value $\rho_{ij} = \sigma_k / \sigma_i \in [0, 1)$.
We clip this as $\rho_{ij} \leftarrow \min(\rho_{ij}, \delta)$ with a scale-invariant parameter $\delta$, which is set as $\delta=\sqrt{0.99}$ for all experiments in this paper. 
We verified the effectiveness of our method by comparing with the method used by \cite{Liao_2019}, as illustrated in Figure~\ref{fig:eps_norm}(a).

Although the clipping above improves numerical stability, 
the norm of the gradient $\bm{\nabla}$ becomes large when $\rho_{ij}$ is close to 1 (Figure~\ref{fig:eps_norm}(b)). 
Since the proposed loss is a convex combination of two losses from the full- and low-rank networks, its gradient is dominated by the one from the low-rank network.
Therefore, we balance these by modifying a parameter $\lambda$ as
\begin{linenomath}
\begin{align}
\lambda \leftarrow \lambda \frac{\| \bm{\nabla}_{f} \|_F}{\| \bm{\nabla} \|_F}, 
\end{align}
\end{linenomath}
where $\bm{\nabla}_{f} = \partial {\mathcal L} ( {\mathcal D}_b, \mathcal W, \Theta ) / \partial \bm{W}$ is the gradient produced by the full-rank network.
This equalizes the gradient norms of the full- and low-rank networks, and thus enables us to effectively adjust the balance with a parameter $\lambda \in [0, 1]$.

%-------------------------------------------------------------------------------------------------------------------------------------------------------------------------------------
\section{Proof of Proposition~\ref{prop:mono}}\label{app:b}
\begin{proof}
As $\bm{y}_{\ell} = \hat{\bm{y}}_{\ell}(R_{\ell})$ by definition, $\| \bm{y}_{\ell} - \hat{\bm{y}}_{\ell}(R_{\ell}) \|^2 = 0$.
To prove 
$\| \bm{y}_{\ell} - \hat{\bm{y}}_{\ell}(1) \|^2 \geq \dots \geq \| \bm{y}_{\ell} - \hat{\bm{y}}_{\ell}(r_{\ell}) \|^2 \geq 
 \| \bm{y}_{\ell} - \hat{\bm{y}}_{\ell}(r_{\ell}+1) \|^2 \geq \dots \geq \| \bm{y}_{\ell} - \hat{\bm{y}}_{\ell}(R_{\ell}) \|^2$, 
we show that $\| \bm{y}_{\ell} - \hat{\bm{y}}_{\ell}(r_{\ell}) \|^2 - \| \bm{y}_{\ell} - \hat{\bm{y}}_{\ell}(r_{\ell}+1) \|^2 \geq 0$ for $1 \leq r_{\ell} \leq R_{\ell}-1$ as 
\begin{linenomath}
\begin{align}\label{eq:prop}
&\| \bm{y}_{\ell} - \hat{\bm{y}}_{\ell}(r_{\ell}) \|^2 - \| \bm{y}_{\ell} - \hat{\bm{y}}_{\ell}(r_{\ell}+1) \|^2 \nonumber \\
&= \| ( \bm{I}_{n_{\ell}} - \bm{V}_{\ell, r_{\ell}} \bm{V}_{\ell, r_{\ell}}^\top ) \bm{y}_{\ell} \|^2 - 
   \| ( \bm{I}_{n_{\ell}} - \bm{V}_{\ell, r_{\ell}+1} \bm{V}_{\ell, r_{\ell}+1}^\top ) \bm{y}_{\ell} \|^2 \nonumber \\
&= \bm{y}_{\ell}^{\top}( \bm{I}_{n_{\ell}} - \bm{V}_{\ell, r_{\ell}} \bm{V}_{\ell, r_{\ell}}^\top ) \bm{y}_{\ell} - 
   \bm{y}_{\ell}^{\top}( \bm{I}_{n_{\ell}} - \bm{V}_{\ell, r_{\ell}+1} \bm{V}_{\ell, r_{\ell}+1}^\top ) \bm{y}_{\ell} \nonumber \\
&= \bm{y}_{\ell}^{\top}( \bm{V}_{\ell, r_{\ell}+1} \bm{V}_{\ell, r_{\ell}+1}^\top - \bm{V}_{\ell, r_{\ell}} \bm{V}_{\ell, r_{\ell}}^\top ) \bm{y}_{\ell} \nonumber \\
&= \bm{y}_{\ell}^{\top}( \bm{v}_{\ell, r_{\ell}+1} \bm{v}_{\ell, r_{\ell}+1}^\top ) \bm{y}_{\ell} \nonumber \\
&= ( \bm{v}_{\ell, r_{\ell}+1}^{\top} \bm{y}_{\ell} )^2 \geq 0.
\end{align}
\end{linenomath}
\noindent
Here, $\bm{V}_{\ell, r_{\ell}} \in \mathbb{R}^{n_{\ell} \times r_{\ell}}$ and $\bm{v}_{\ell, r_{\ell}+1} \in \mathbb{R}^{n_{\ell}}$ are the slicing of $\bm{V}_{\ell}[:, :r_{\ell}]$ 
and the $(r_{\ell}+1)$-th column vector in $\bm{V}_{\ell}$, respectively. 
$\bm{I}_{n_{\ell}}$ indicates the identity matrix with size $n_{\ell} \times n_{\ell}$. \\
\end{proof}

%-------------------------------------------------------------------------------------------------------------------------------------------------------------------------------------
\section{Proof of Proposition~\ref{prop:bound_l2}}\label{app:c}
Let $\hat{\bm{y}}_{\ell} = \widetilde{\bm{W}}_{\ell}^{\top} \bm{x}_{\ell}$ be an approximated output with a low-rank matrix $\widetilde{\bm{W}}_{\ell}$ in the layer $\ell$ 
and let $\widetilde{\bm{y}}_{\ell} = \widetilde{\bm{W}}_{\ell}^{\top} \widetilde{\bm{x}}_{\ell}$ be a fully approximated output up to the layer $\ell$ 
with $\widetilde{\bm{x}}_{\ell} = \phi(\widetilde{\bm{y}}_{\ell-1})$.
Let $p_k=\frac{\exp(y_{L, k})}{\sum_{j=1}^K\exp(y_{L, j})}$ and $\tilde{p}_k=\frac{\exp(\widetilde{y}_{L, k})}{\sum_{j=1}^K\exp(\widetilde{y}_{L, j})}$ for $1 \leq k \leq K$ 
be $K$-class softmax outputs from the full- and low-rank networks, respectively.
\begin{proof} 
The squared error between $\bm{y}_{\ell}$ and $\widetilde{\bm{y}}_{\ell}$ can be evaluated as 
\begin{linenomath}
\begin{align*}
\| \bm{y}_{\ell} - \widetilde{\bm{y}}_{\ell} \|^2 \
&= \left\| \bm{W}_{\ell}^{\top} \bm{x}_{\ell} - \widetilde{\bm{W}}_{\ell}^{\top} \widetilde{\bm{x}}_{\ell} \right\|^2 \\
&= \left\| \bm{W}_{\ell}^{\top} \bm{x}_{\ell} - \widetilde{\bm{W}}_{\ell}^{\top} \widetilde{\bm{x}}_{\ell}
   + \widetilde{\bm{W}}_{\ell}^{\top} \bm{x}_{\ell} - \widetilde{\bm{W}}_{\ell}^{\top} \bm{x}_{\ell} \right\|^2 \\
&= \left\| (\bm{W}_{\ell} - \widetilde{\bm{W}}_{\ell})^{\top} \bm{x}_{\ell} + \widetilde{\bm{W}}_{\ell}^{\top} (\bm{x}_{\ell} - \widetilde{\bm{x}}_{\ell}) \right\|^2 \\
&= \left\| (\bm{W}_{\ell} - \widetilde{\bm{W}}_{\ell})^{\top} \bm{x}_{\ell} \right\|^2 
 + \left\| \widetilde{\bm{W}}_{\ell}^{\top} (\bm{x}_{\ell} - \widetilde{\bm{x}}_{\ell}) \right\|^2 \\
&\quad+ 2\bm{x}_{\ell}^{\top} \underbrace{(\bm{W}_{\ell} - \widetilde{\bm{W}}_{\ell}) \widetilde{\bm{W}}_{\ell}^{\top}}_{\bm{O}} (\bm{x}_{\ell} - \widetilde{\bm{x}}_{\ell}) \\
&= \left\| (\bm{W}_{\ell} - \widetilde{\bm{W}}_{\ell})^{\top} \bm{x}_{\ell} \right\|^2 
 + \left\| \widetilde{\bm{W}}_{\ell}^{\top} (\bm{x}_{\ell} - \widetilde{\bm{x}}_{\ell}) \right\|^2 \\
&= \left\| \bm{y}_{\ell} - \hat{\bm{y}}_{\ell} \right\|^2 
 + \left\| \widetilde{\bm{W}}_{\ell}^{\top} (\bm{x}_{\ell} - \widetilde{\bm{x}}_{\ell}) \right\|^2 \\
&\leq \left\| \bm{y}_{\ell} - \hat{\bm{y}}_{\ell} \right\|^2 + \| \widetilde{\bm{W}}_{\ell} \|_2^2 \left\| \bm{x}_{\ell} - \widetilde{\bm{x}}_{\ell} \right\|^2 \\
&= \left\| \bm{y}_{\ell} - \hat{\bm{y}}_{\ell} \right\|^2 + \| \bm{W}_{\ell} \|_2^2 \left\| \bm{x}_{\ell} - \widetilde{\bm{x}}_{\ell} \right\|^2, 
\end{align*}
\end{linenomath}
where we assume that $\| \widetilde{\bm{W}}_{\ell} \|_2^2 = \| \bm{W}_{\ell} \|_2^2$ because basis corresponding to the maximum singular value remains unchanged.
For 1-Lipschitz activation function (e.g., ReLU), we can evaluate as 
$
\left\| \bm{x}_{\ell} - \widetilde{\bm{x}}_{\ell} \right\| 
= \left\| \phi(\bm{y}_{\ell-1}) - \phi(\widetilde{\bm{y}}_{\ell-1}) \right\| 
\leq \left\| \bm{y}_{\ell-1} - \widetilde{\bm{y}}_{\ell-1} \right\|
$, and thus obtain 
\begin{linenomath}
\begin{align*}
\| \bm{y}_{\ell} - \widetilde{\bm{y}}_{\ell} \|^2 
\leq \left\| \bm{y}_{\ell} - \hat{\bm{y}}_{\ell} \right\|^2 + \| \bm{W}_{\ell} \|_2^2 \left\| \bm{y}_{\ell-1} - \widetilde{\bm{y}}_{\ell-1} \right\|^2.
\end{align*}
\end{linenomath}
Evaluating the above inequality recursively for $\ell=2 \sim L$, 
the squared error between outputs in the last layer (i.e., logits) can be bounded as 
\begin{linenomath}
\begin{align}\label{eq:bound_logit}
&\|\bm{y}_L - \widetilde{\bm{y}}_L\|^2 \leq \nonumber \\
&\|\bm{y}_L - \hat{\bm{y}}_L\|^2 + \sum_{\ell=1}^{L-1} \left\{ \|\bm{y}_{\ell}-\hat{\bm{y}}_{\ell}\|^2 \prod_{j=\ell+1}^{L} \|\bm{W}_{j}\|_2^2 \right\}.
\end{align}
\end{linenomath}

Next, we show that 
${\rm KL}(\bm{p}||\widetilde{\bm{p}}) \leq \sqrt{2} \left\|\bm{y}_{L} - \widetilde{\bm{y}}_{L} \right\|$.
According to the nonnegativity of KL-divergence, 
\begin{linenomath}
\begin{align*}
&{\rm KL}(\bm{p}||\widetilde{\bm{p}}) \leq {\rm KL}(\bm{p}||\widetilde{\bm{p}}) + {\rm KL}(\widetilde{\bm{p}}||\bm{p}) \\
&= \sum_{k=1}^K \left( p_k\log{p_k} - p_k\log{\widetilde{p}_k} + \widetilde{p}_k\log{\widetilde{p}_k} - \widetilde{p}_k\log{p_k} \right) \\
&= \sum_{k=1}^K \left( p_k - \widetilde{p}_k \right) \left(\log{p_k} - \log{\widetilde{p}_k} \right) \\
&= \sum_{k=1}^K \left( p_k - \widetilde{p}_k \right) \left(y_{L, k} - \widetilde{y}_{L, k}
 + \log{\frac{\sum_{j=1}^K\exp(\widetilde{y}_{L, j})}{\sum_{j=1}^K\exp(y_{L, j})}} \right) \\
&= \sum_{k=1}^K \left( p_k - \widetilde{p}_k \right) \left(y_{L, k} - \widetilde{y}_{L, k} \right) \\
&\quad + \underbrace{\left(\sum_{k=1}^K \left( p_k - \widetilde{p}_k \right)\right)}_{0} \log{\frac{\sum_{j=1}^K\exp(\widetilde{y}_{L, j})}{\sum_{j=1}^K\exp(y_{L, j})}} \\
&= \sum_{k=1}^K \left( p_k - \widetilde{p}_k \right) \left(y_{L, k} - \widetilde{y}_{L, k} \right) \\
&\leq \sqrt{\left(\sum_{k=1}^K \left( p_k - \widetilde{p}_k \right)^2 \right) 
         \left(\sum_{k=1}^K \left(y_{L, k} - \widetilde{y}_{L, k} \right)^2 \right)}.
\end{align*}
\end{linenomath}
Moreover, noting that $0 \leq p_k \leq 1$, $0 \leq \widetilde{p}_k \leq 1$, $\sum_{k=1}^K p_k = \sum_{k=1}^K \widetilde{p}_k = 1$, 
we obtain
\begin{linenomath}
\begin{align*}
\sum_{k=1}^K \left( p_k - \widetilde{p}_k \right)^2
&= \sum_{k=1}^K \left( p_k \right)^2 - 2\sum_{k=1}^K p_k \widetilde{p}_k + \sum_{k=1}^K \left( \widetilde{p}_k \right)^2 \\
&\leq \sum_{k=1}^K p_k - 2\sum_{k=1}^K p_k \widetilde{p}_k + \sum_{k=1}^K \widetilde{p}_k \\
&= 2 \left( 1 - \sum_{k=1}^K p_k \widetilde{p}_k \right) \\
&\leq 2.
\end{align*}
\end{linenomath}
Therefore, we have 
\begin{linenomath}
\begin{align*}
{\rm KL}(\bm{p}||\widetilde{\bm{p}}) 
&\leq \sqrt{2 \left(\sum_{k=1}^K \left(y_{L, k} - \widetilde{y}_{L, k} \right)^2 \right)} \nonumber \\
&= \sqrt{2} \left\|\bm{y}_{L} - \widetilde{\bm{y}}_{L} \right\|.
\end{align*}
\end{linenomath}
Combining the above with inequality~(\ref{eq:bound_logit}), we finally derive 
\begin{linenomath}
\begin{align*}
&{\rm KL}(\bm{p}||\widetilde{\bm{p}}) 
\leq \nonumber \\
&\sqrt{2 \left( \|\bm{y}_L - \hat{\bm{y}}_L\|^2 + \sum_{\ell=1}^{L-1} \left\{ \|\bm{y}_{\ell}-\hat{\bm{y}}_{\ell}\|^2 \prod_{j=\ell+1}^{L} \|\bm{W}_{j}\|_2^2 \right\} \right)}.
\end{align*}
\end{linenomath}
\end{proof}

%-------------------------------------------------------------------------------------------------------------------------------------------------------------------------------------
\section{Additional Details of Experiments}\label{app:d}
\begin{table}[tb]
\begin{center}
	\small
	\begin{tabularx}{\linewidth}{l|X}
	\bhline{1pt}
	Preprocess				&	Per-channel standardization:													\\
							&	\hspace{7.5pt} CIFAR-10:														\\
							&	\hspace{15pt} $\texttt{mean}=(0.4914, 0.4822, 0.4465)$							\\
							&	\hspace{15pt} $\texttt{std.}=(0.2470, 0.2435, 0.2616)$							\\
							&	\hspace{7.5pt} CIFAR-100:														\\
							&	\hspace{15pt} $\texttt{mean}=(0.5071, 0.4865, 0.4409)$							\\
							&	\hspace{15pt} $\texttt{std.}=(0.2673, 0.2564, 0.2762)$							\\
	\hline
	Augmentation			&	\begin{itemize}
									\item Random cropping 32 $\times$ 32 after zero-padding 4 pixels in each corner
									\item Random horizontal flipping ($p=0.5$)
								\end{itemize}																	\\
	\hline
	Batch size				&	$B=128$																			\\
	\hline
	Epoch					&	200																				\\
	\hline
	Optimizer				&	SGD with Nesterov momentum ($\mu=0.9$)											\\
	\hline
	Learning rate			&	Initially 0.1, decaying by rate 0.2 at 60, 120, and 160 epochs					\\
	\hline
	$L_2$ regularization	&	$\eta=0.0005$																	\\
	\hline
	Initializer				&	He-Normal~\cite{He_2015} for weights, 0 for biases								\\ 
	\hline
	BN						&	$\epsilon=1.0 \times 10^{-5}, \texttt{momentum}=0.9$, initializing $\gamma=1$ and $\beta=0$\\
	\hline
	GPU(s)					&	1																				\\
	\bhline{1pt}
	\end{tabularx}
	\caption{Baseline setups for the CIFAR-10/100 datasets.}
	\label{tbl:setup_cifar}
\end{center}
\end{table}
\begin{table}[tb]
\begin{center}
	\small
	\begin{tabularx}{\linewidth}{l|X}
	\bhline{1pt}
	Preprocess				&	Per-channel standardization:													\\
							&	\hspace{15pt} $\texttt{mean}=(0.485, 0.456, 0.406)$								\\
							&	\hspace{15pt} $\texttt{std.}=(0.229, 0.224, 0.225)$								\\
	\hline
  	Augmentation			&	\begin{itemize}
  									\item Random resized cropping $224 \times 224$ with scale range (0.08, 1) and aspect ratio range (3/4, 4/3)
 									\item ColorJitter with brightness 0.4, contrast 0.4, saturation 0.4, and hue 0
									\item PCA-lighting with $\texttt{std.}=0.1$
									\item Random horizontal flipping ($p=0.5$)
								\end{itemize}																	\\
	\hline
	Batch size				&	$B=256$																			\\
	\hline
	Epoch					&	100																				\\
	\hline
	Optimizer				&	SGD with Nesterov momentum ($\mu=0.9$)											\\
	\hline
	Learning rate			&	Initially 0.1, decaying by rate 0.1 at 30, 60, and 90 epochs					\\
	\hline
	$L_2$ regularization	&	$\eta=0.0001$																	\\
	\hline
	Initializer				&	He-Normal~\cite{He_2015} for weights, 0 for biases								\\
	\hline
	BN						&	$\epsilon=1.0 \times 10^{-5}, \texttt{momentum}=0.9$, initializing $\gamma=1$ and $\beta=0$\\
	\hline
	GPU(s)					&	8																				\\
	\bhline{1pt}
	\end{tabularx}
	\caption{Baseline setups for the ImageNet dataset.}
	\label{tbl:setup_imagenet}
\end{center}
\end{table}
\begin{table}[tb]
\begin{center}
	\small
	\begin{tabularx}{\linewidth}{l|X}
	\bhline{1pt}
	Model			&	NVIDIA DGX-1								\\
	\hline
	CPU				&	Dual 20-Core Intel Xeon E5-2698 v4 2.2 GHz	\\
	\hline
	GPU				&	8$\times$ Tesla V100						\\
	\hline
	RAM				&	512 GB 2,133 MHz DDR4						\\
	\hline
	OS				&	Ubuntu 18.04 LTS							\\
	\hline
	Framework		&	Tensorflow 1.15.0	\\
	\bhline{1pt}
	\end{tabularx}
	\caption{Computing infrastructure used for running experiments.}
	\label{tbl:setup_machine}
\end{center}
\end{table}

In all experiments, we follow the same baseline setup for the CIFAR datasets as used by \cite{Zagoruyko_2016}, 
and the setup of \cite{Yu_2019_1} for the ImageNet dataset. 
Except for the parameters specific to each method, we use the same setups summarized in Table~\ref{tbl:setup_cifar} and \ref{tbl:setup_imagenet}, respectively for each dataset. 
For validation on the ImageNet dataset, each image is resized so that a short side has a length of 256, then the center of which is cropped with a size of $224 \times 224$.
The computing infrastructure used for running experiments is also shown in Table~\ref{tbl:setup_machine}.

In the comparison experiment with retraining-based methods, we perform fine-tuning after decomposition, 
in which training parameters are selected through a grid search. 
In particular, learning rate and weight decay are selected from among $\{0.0005, 0.001, 0.005, 0.01\}$ and $\{0.00005, 0.0001, 0.0005, 0.001\}$ for the CIFAR datasets, 
and among $\{0.001, 0.01\}$ and $\{0.0001, 0.001\}$ for the ImageNet dataset. 
The learning rate scheduler is selected from either \texttt{step decay} at 5 and 10 epochs or \texttt{polynomial decay} with a last learning rate of 0 for the CIFAR datasets.
The decay rate of the former and the polynomial power of the latter one are respectively selected from among $\{0.5, 0.2, 0.1\}$ and $\{1, 1.5, 2\}$.
For the ImageNet dataset, the learning rate scheduler is \texttt{step decay} at 5 and 10 epochs and its decay rate is selected from among $\{0.2, 0.1\}$.
We fix the number of epochs to 15, as suggested by \cite{Kim_2016}.
The best results from Tucker decomposition are achieved by \texttt{polynomial decay} for the CIFAR datasets, and its parameters (learning rate, weight decay, and polynomial power) 
are (0.01, 0.001, 1.5) for the CIFAR-10 dataset and (0.01, 0.001, 1.5) for the CIFAR-100 dataset.
For the ImageNet dataset, the best parameters are a learning rate of 0.01, a weight decay of 0.0001, and a step decay rate of 0.2. 

%-------------------------------------------------------------------------------------------------------------------------------------------------------------------------------------
\section{An Empirical Analysis for Lipschitz Continuity}\label{app:e}
\begin{figure*}[t]
	\centerline{\includegraphics[width=\linewidth]{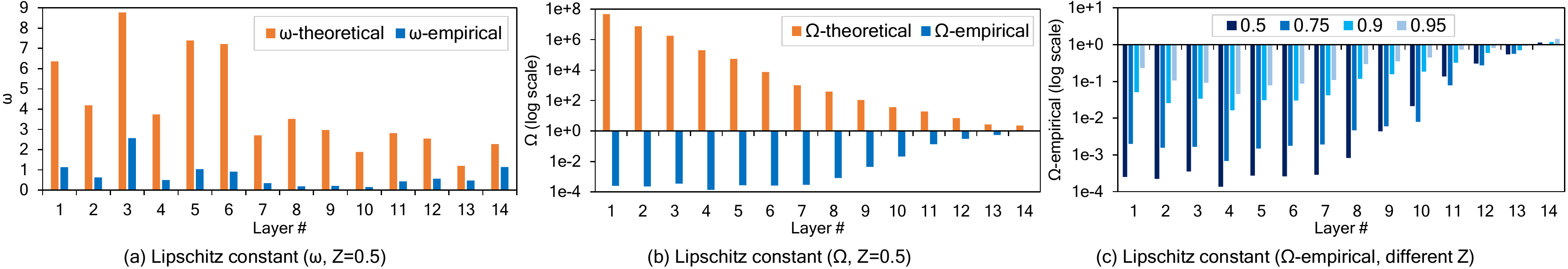}}
	\caption{
		Lipschitz constants for VGG-15 trained on CIFAR-10. 
		(a) \texttt{$\omega$-theoretical} and \texttt{$\omega$-empirical} indicate $\omega_{\ell}$ and $\widehat{\omega}_{\ell}$, and 
		(b) \texttt{$\Omega$-theoretical} and \texttt{$\Omega$-empirical} indicate $\Omega_{\ell}$ and $\widehat{\Omega}_{\ell}$, respectively. 
		Those are evaluated for $\ell \in \{1, \dots, 14\}$ in the compressed model with the rank ratio $Z=0.5$. 
		(c) $\widehat{\Omega}_{\ell}$ is evaluated for $\ell \in \{1, \dots, 14\}$ in the four compressed models with different rank ratios $Z \in \{0.5, 0.75, 0.9, 0.95\}$.
	}
	\label{fig:lipschitz}
\end{figure*}
\begin{figure*}[h]
	\centerline{\includegraphics[width=\linewidth]{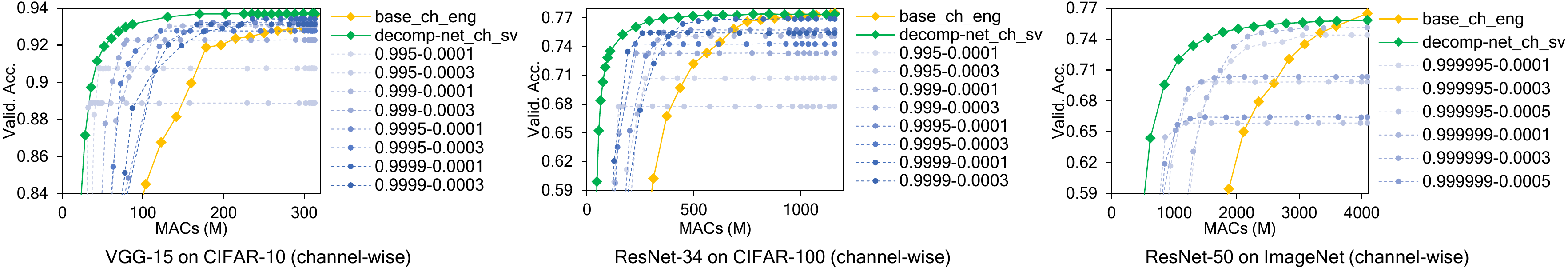}}
	\caption{
		Additional comparison results.
		Dashed lines are the results from TRP~\protect\cite{Xu_2019}, which are individually trained with different parameters ($e-\tau$).
		\texttt{base\_ch\_eng} is a baselline with channel-wise decomposition, in which normal learning is adopted instead of TRP.
	}
	\label{fig:comparison_sm1}
\end{figure*}
As described in section~\ref{subsec:theory}, 
an upper bound of KL divergence between softmax outputs from the full- and low-rank networks 
is characterized by the layer-wise error $\| \bm{y}_{\ell} - \hat{\bm{y}}_{\ell} \|^2$. 
Since the monotonicity of the error holds for each layer, 
it can be expected that performance of the entire network changes along with changing the rank in each layer. 
However, the layer-wise error in the upper bound is multiplied by the product of Lipschitz constants $\prod_{j=\ell+1}^{L} \|\bm{W}_{j}\|_2^2$ for $\ell \in \{1, \dots, L-1\}$, 
and that is exponentially amplified when $\|\bm{W}_{j}\|_2 > 1$. 
This leads the upper bound to become loose. 
In order to validate that the upper bound is not trivial in practice, 
we experimentally show that the Lipschitz constant of the network is much smaller than the theoretical one, as reported by \cite{Arora_2018,Wei_2019}.

In deriving the upper bound, we evaluate the approximation error in a layer $\ell$ as
\begin{linenomath}
\begin{align*}
\| \bm{y}_{\ell} - \widetilde{\bm{y}}_{\ell} \|^2 \
&= \left\| \bm{y}_{\ell} - \hat{\bm{y}}_{\ell} \right\|^2 
 + \left\| \widetilde{\bm{W}}_{\ell}^{\top} (\phi(\bm{y}_{\ell-1}) - \phi(\widetilde{\bm{y}}_{\ell-1})) \right\|^2 \\
&\leq \left\| \bm{y}_{\ell} - \hat{\bm{y}}_{\ell} \right\|^2 + \| \widetilde{\bm{W}}_{\ell} \|_2^2 \left\| \bm{y}_{\ell-1} - \widetilde{\bm{y}}_{\ell-1} \right\|^2, 
\end{align*}
\end{linenomath}
based on the relation that 
\begin{linenomath}
\begin{align*}
\left\| \widetilde{\bm{W}}_{\ell}^{\top} (\phi(\bm{y}_{\ell-1}) - \phi(\widetilde{\bm{y}}_{\ell-1})) \right\|^2
&\leq \| \widetilde{\bm{W}}_{\ell} \|_2^2 \left\| \bm{y}_{\ell-1} - \widetilde{\bm{y}}_{\ell-1} \right\|^2. 
\end{align*}
\end{linenomath}
In other words, the Lipschitz constant in a layer $\ell$ is $\omega_{\ell} = \| \widetilde{\bm{W}}_{\ell} \|_2~(=\| \bm{W}_{\ell} \|_2)$, 
which can be interpreted as the maximum rate of change in the approximation error. 
Here, we evaluate its empirical counterpart by using the training dataset of $N$ samples as
\begin{linenomath}
\begin{align*}
\widehat{\omega}_{\ell} = \max_{n \in \{1,\ldots, N\}} \frac{\left\| \widetilde{\bm{W}}_{\ell}^{\top} (\phi(\bm{y}_{\ell-1, n}) - \phi(\widetilde{\bm{y}}_{\ell-1, n})) \right\|}{\left\| \bm{y}_{\ell-1, n} - \widetilde{\bm{y}}_{\ell-1, n} \right\|},
\end{align*}
\end{linenomath}
for $\bm{y}_{\ell-1,n} \neq \widetilde{\bm{y}}_{\ell-1,n}$. 
Then, we compare the theoretical and empirical values of the Lipschitz constant of the network: 
\begin{linenomath}
\begin{align*}
\Omega_{\ell} = \prod_{j=\ell+1}^{L} \omega_{j}, \ \ 
\widehat{\Omega}_{\ell} = \prod_{j=\ell+1}^{L} \widehat{\omega}_{j}, 
\end{align*}
\end{linenomath}
for $\ell \in \{1, \dots, L-1\}$.

We evaluate the Lipschitz constants for VGG-15 on the CIFAR-10 dataset with the setups described in Table~\ref{tbl:setup_cifar}. 
In this case, $L=15$ and 
$\omega_{\ell}=\| \bm{W}_{\ell} \|_2 \left\lceil \frac{K_{\ell}}{S_{\ell}} \right\rceil$ for the convolutional layers, 
where $\lceil \cdot \rceil$ indicates the ceiling function, and $K_{\ell}$ and $S_{\ell}$ are the kernel size and stride, respectively. 
In order to evaluate the empirical Lipschitz constant, the trained model is compressed in terms of the rank ratio $Z$. 
Figure~\ref{fig:lipschitz}(a) shows $\omega_{\ell}$ and $\widehat{\omega}_{\ell}$ for each layer in the model with $Z=0.5$, 
likewise Figure~\ref{fig:lipschitz}(b) shows $\Omega_{\ell}$ and $\widehat{\Omega}_{\ell}$ in the same model.
We can see that $\omega_{\ell} > 1$ for all layers in Figure~\ref{fig:lipschitz}(a), 
and therefore $\Omega_{\ell}$ becomes exponentially larger for earlier layers as shown in Figure~\ref{fig:lipschitz}(b).
On the other hand, it can be observed that the empirical values ($\widehat{\omega}_{\ell}$) are much smaller than the theoretical ones in Figure~\ref{fig:lipschitz}(a).
Since there are some layers in which $\widehat{\omega}_{\ell}$ are less than $1$, 
$\widehat{\Omega}_{\ell}$ does not become larger for earlier layers, but rather become smaller than for later layers, as shown in Figure~\ref{fig:lipschitz}(b). 
We consider that this behavior is related to the noise stability of trained DNNs~\cite{Arora_2018}, 
which is the property that the error injected in an earlier layer tends to be attenuated as it propagates through layers. 
To investigate the property, we additionally compare $\widehat{\Omega}_{\ell}$ in the four compressed models with different rank ratios $Z \in \{0.5, 0.75, 0.9, 0.95\}$. 
Figure~\ref{fig:lipschitz}(c) shows that the model with smaller $Z$ tends to have smaller $\widehat{\Omega}_{\ell}$. 
Therefore, the error in an earlier layer is more likely to be attenuated in the lower rank models, which is also correlated with the analysis by \cite{Arora_2018}. 

In addition, we can observe in Figure~\ref{fig:lipschitz}(b) that there is no much difference in $\widehat{\Omega}_{\ell}$ of up to the middle layer (\#7), 
but $\widehat{\Omega}_{\ell}$ is larger in the later layer. 
This implies that the approximation error incurred in the later layer has a greater influence on the upper bound for the entire network.
Therefore, we can consider to adopt a higher importance to the later layer in rank selection, so that the upper bound is minimized more efficiently.
We leave consideration of this for future works.
%-------------------------------------------------------------------------------------------------------------------------------------------------------------------------------------
\end{document}